\RequirePackage{fix-cm}
\RequirePackage{fixltx2e}
\documentclass[twoside,english,aip, jcp,floatfix]{revtex4-1}
\usepackage[T1]{fontenc}
\usepackage[latin9]{inputenc}
\usepackage[a4paper]{geometry}
\usepackage{xcolor}
\usepackage{babel}
\usepackage{amsmath}
\usepackage{amssymb}
\usepackage{graphicx}
\usepackage[unicode=true,pdfusetitle,
 bookmarks=false,
 breaklinks=true,pdfborder={0 0 0},pdfborderstyle={},backref=false,colorlinks=true]
 {hyperref}
\hypersetup{
 citecolor = blue, linkcolor = blue,  urlcolor  = blue}

\makeatletter

\providecommand{\tabularnewline}{\\}

\usepackage{orcidlink}

\makeatother

\begin{document}

\title{{\Large{}Properties of the ENCE and other MAD-based calibration metrics}}

\author{Pascal PERNOT \orcidlink{0000-0001-8586-6222}}

\affiliation{Institut de Chimie Physique, UMR8000 CNRS,~\\
Université Paris-Saclay, 91405 Orsay, France}
\email{pascal.pernot@cnrs.fr}

\begin{abstract}
\noindent The Expected Normalized Calibration Error (ENCE) is a popular
calibration statistic used in Machine Learning to assess the quality
of prediction uncertainties for regression problems. Estimation of
the ENCE is based on the binning of calibration data. In this short
note, I illustrate an annoying property of the ENCE, i.e. its proportionality
to the square root of the number of bins for well calibrated or nearly
calibrated datasets. A similar behavior affects the calibration error
based on the variance of $z$-scores (ZVE), and in both cases this
property is a consequence of the use of a Mean Absolute Deviation
(MAD) statistic to estimate calibration errors. Hence, the question
arises of which number of bins to choose for a reliable estimation
of calibration error statistics. A solution is proposed to infer ENCE
and ZVE values that do not depend on the number of bins for datasets
assumed to be calibrated, providing simultaneously a statistical calibration
test. It is also shown that the ZVE is less sensitive than the ENCE
to outstanding errors or uncertainties.
\end{abstract}
\maketitle

\subsection{Definition of the ENCE}

\noindent Various forms of Mean Absolute Deviation (MAD) statistics
have been proposed to quantify the calibration error in Machine Learning
(ML) classification\citep{Guo2017} or regression problems\citep{Laves2020,Levi2020,Levi2022}
with Uncertainty Quantification (UQ). Among them, the Expected Normalized
Calibration Error (ENCE) has been introduced recently by Levi \emph{et
al.}\citep{Levi2020,Levi2022}, and used in several ML-UQ studies
since\citep{Scalia2020,Wang2021,Busk2022,Vazquez-Salazar2022,Frenkel2023}.
I focus here on the ENCE, but is variants such as the Uncertainty
Calibration Error (UCE or ECE) are also concerned.

Let us consider a set of prediction errors and uncertainties $\left\{ E_{i},u_{i}\right\} _{i=1}^{M}$
arising from an ML-UQ regression problem. \emph{Average} calibration
can be tested by comparing the mean variance (MV) estimated from the
uncertainties
\begin{equation}
\mathrm{MV}=\frac{1}{M}\sum_{i=1}^{M}u_{i}^{2}
\end{equation}
to the variance of the errors, or in the hypothesis of unbiased errors
($\overline{E}\simeq0$), to the mean squared errors (MSE) 
\begin{equation}
\mathrm{MSE}=\frac{1}{M}\sum_{i=1}^{M}E_{i}^{2}
\end{equation}

\noindent Average calibration is know to be unreliable to characterize
the properties of prediction uncertainties, notably because an average
of overestimated and underestimated uncertainties might still provide
a MV value in agreement with the MSE.\citep{Pernot2022a,Pernot2022b}
To avoid this, it was proposed to use \emph{conditional} calibration,
in which calibration is tested \emph{locally} across the range of
uncertainty values.\citep{Levi2020,Levi2022,Pernot2023_Arxiv} Conditional
calibration with respect to uncertainty is also called \emph{consistency},
to distinguish if from conditional calibration with respect to \emph{input
features}, which is called \emph{adaptivity}.\citep{Pernot2023_Arxiv,Angelopoulos2021}

In order to test consistency, one procedure is to compare the MV to
the MSE using \emph{binned} data. For this, the dataset is ordered
by increasing uncertainty values and split into $N$ bins $B_{1},\ldots,B_{N}$,
which is expected to reduce the possibility of compensation between
over- and under-estimated uncertainties when the bin size is chosen
small enough. The \emph{local} calibration within bin $i$ can be
assessed by checking that the mean variance
\begin{equation}
\mathrm{MV}_{i}=\frac{1}{|B_{i}|}\sum_{j\in B_{i}}u_{j}^{2}
\end{equation}
is equal to mean squared error
\begin{equation}
\mathrm{MSE}_{i}=\frac{1}{|B_{i}|}\sum_{j\in B_{i}}E_{j}^{2}
\end{equation}
where $|B_{i}|$ is the size of bin $i$.

\emph{Reliability diagrams} are obtained by plotting $\mathrm{MSE}_{i}$
vs $\mathrm{MV}_{i}$ and by comparing the points to the identity
line.\citep{Levi2020,Levi2022} In order to quantify the amplitude
of the differences between the error and uncertainty statistics, one
estimates the ENCE metric, as the relative MAD between $\mathrm{MV}_{i}^{1/2}$
and $\mathrm{MSE}_{i}^{1/2}$
\begin{equation}
\mathrm{ENCE}=\frac{1}{N}\sum_{i=1}^{N}\frac{|\mathrm{MV}_{i}^{1/2}-\mathrm{MSE}_{i}^{1/2}|}{\mathrm{MV}_{i}^{1/2}}
\end{equation}
The ENCE is typically reported as a percentage, and used to compare
the performances of various calibration procedures on a given dataset.

\subsection{Impact of the binning scheme on ENCE estimation}

\noindent One can observe in the literature a diversity of binning
strategies to estimate the ENCE. Some authors use a small number of
bins (typically 10-15)\citep{Laves2020,Levi2022,Busk2022,Palmer2022},
whereas other prefer much larger numbers \citep{Scalia2020}. There
is also a choice between equal-size bins and equal-width bins, but
for the present study I focus only on the former strategy. 

Several considerations can help to define a proper number of bins.
The use of a large number of bins seems more in agreement with the
approach of conditional calibration, but smaller numbers of bins can
be expected to provide more accurate statistics. It is indeed important
that the bin size is kept large enough to preserve testing power when
the bins are used to estimate test statistics\citep{Pernot2022a}.
For equal-size bins, the distribution of binned values is transformed
to a uniform distribution, and entropy-based arguments for histograms
provide the optimal number of bins as the square root of the sample
size ($N=M^{1/2}$).\citep{Watts2022} This tends to favor rather
large bin numbers for datasets with more than 1000 points. 

As there is presently no consensus on the optimal binning strategy
for the estimation of calibration errors, it is interesting to check
how the choice of $N$ influences ENCE values. In this perspective,
let us first consider a slightly simplified problem to illustrate
the estimation issues (a fully analytical example loosely related
to the ENCE is also presented in Appendix\,\ref{subsec:MAD-of-binned}).
Assuming an homoscedastic dataset ($u_{i}=u=c^{te}$), one can write
\begin{equation}
\mathrm{ENCE}=\frac{1}{Nu}\sum_{i=1}^{N}|\mathrm{MSE}_{i}^{1/2}-u|
\end{equation}
For \emph{normally} distributed errors ($E_{i}\sim N(0,1)$), $\mathrm{MSE}_{i}^{1/2}$
has a \emph{chi distribution} with $k$ degrees of freedom, $\chi_{k}$,
where $k$ is the size of a bin ($k=\left\lfloor M/N\right\rfloor $).
Due to the absolute value, the ENCE is thus based on the sum of shifted-folded
$\chi_{k}$ distributions, the properties of which have to be studied
by simulation. 

\begin{figure}[t]
\noindent \begin{centering}
\includegraphics[width=0.45\textwidth]{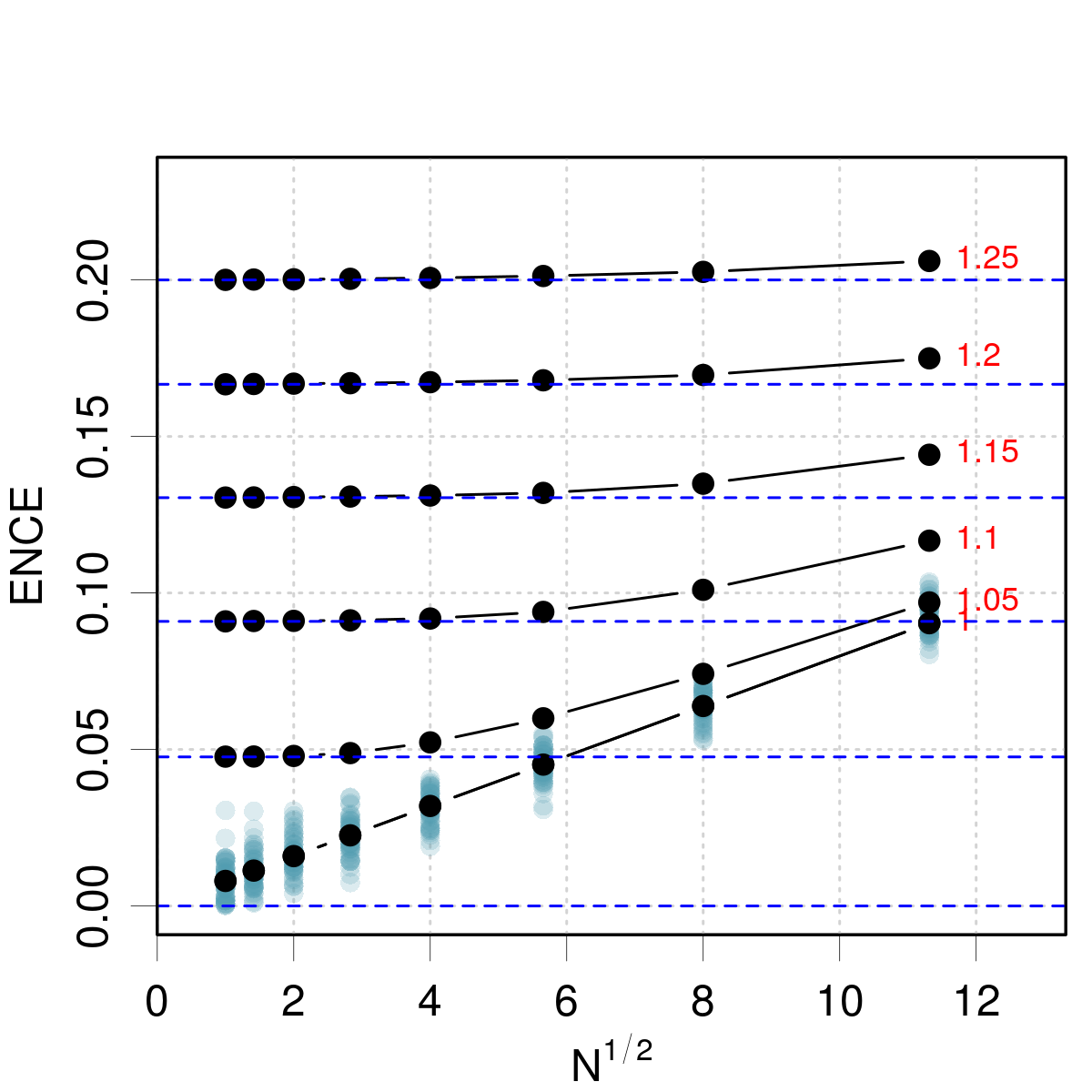}
\par\end{centering}
\caption{\label{fig:Bin-size-dependence-of-ENCE}Dependence of the ENCE on
the square root of the number of bins $N^{1/2}$ for several miscalibration
factors (red numbers). The dataset size is $M=5000$. For each bin
size $k=M/N$, the ENCE is estimated by Monte Carlo: $N$ samples
$X$ of $10^{5}$ points of a $\chi_{k}$ variable are drawn; for
each bin the sample mean of $|X-u|/u$ is estimated; finally the ENCE
is obtained by averaging over the bins. The blue dots represent 50
realizations of the estimation of ENCE from random $E\sim N(0,1)$
datasets analyzed by the \texttt{ErrViewLib::plotRelDiag} function.}
\end{figure}

Fig.\,\ref{fig:Bin-size-dependence-of-ENCE} reports the dependence
of the ENCE for a well calibrated dataset $(u=1)$ and several levels
of miscalibration, with $u$ varying between 1.05 and 1.25.
For the calibrated dataset ($u=1$) the ENCE estimated using the $\chi_{k}$
distribution approach is proportional to $N^{1/2}$. To validate this
result, a Monte Carlo simulation of ENCE values from random samples
of $E$ was performed and plotted as blue dots. The curve using the
$\chi_{k}$ distributions represents perfectly the average location
of these points for each value of $N$, and as it is much less computer
intensive, it was used for the other values of $u$. For uncalibrated
datasets ($u\ne1$), one observes a transient behavior between a constant
value (giving the correct relative error) for small values of $N$
and a linear dependence with $N^{1/2}$ as $N$ increases. The constant
part becomes predominant as $u$ increases.

This transient behavior of the ENCE is due to the properties of the
MAD, which measures a bias when all deviations $\mathrm{MSE}_{i}^{1/2}-u$
are of the same sign (the ENCE is independent of $N$) and a dispersion
when $\mathrm{MSE}_{i}^{1/2}\simeq u$ (the ENCE depends on the uncertainty
of the binned statistic, see Appendix\,\ref{subsec:MAD-of-binned}).\citep{Pernot2018}

The ENCE seems thus well adapted to characterize a lack of calibration,
and using small numbers of bins as often done in the literature seems
to provide a reliable estimation of the relative calibration error.
However, one is facing a problem when testing a calibrated dataset,
for which any reasonable value of $N$ gives a non-zero and different
estimation of the ENCE. In this case, the ENCE is dominated by the
statistical fluctuations of the binned values. For our model example,
using $N=36$ bins would give a 5\,\% ENCE. How does this inform
us on the calibration of this dataset ?

One might be tempted to use the $N^{1/2}$ law as a baseline to test
calibration, but the proportionality constant is sensitive to the
errors and uncertainties distribution, and therefore dataset-dependent
(see Table\,\ref{tab:Coefficients-of-the}). A better solution could
be to perform a linear regression of ENCE values obtained for several
values of $N$ versus $N^{1/2}$ and compare the intercept to 0 (the
confidence interval for the intercept of a perfectly calibrated dataset
should include 0). This method is tested below.

\subsection{Definition of the Z-Variance Error (ZVE)}

\noindent An alternative way to test calibration\citep{Pernot2022b}
is to check that the variance of $z$-scores is equal to 1
\begin{equation}
\mathrm{Var}\left(Z=E/u\right)=1\label{eq:LZV}
\end{equation}
which extends directly to individual bins
\begin{equation}
v_{i}=\mathrm{Var}\left(\left\{ Z_{j}=E_{j}/u_{j}\right\} _{j\in B_{i}}\right)=1\label{eq:LZV-1}
\end{equation}
as used in the Local Z-Variance (LZV) analysis.\citep{Pernot2022b}
This approach is preferable to the MV/MSE approach as it accounts
explicitly for the pairing of errors and uncertainties and is less
susceptible to fortuitous compensations.

Following the same ideas as for the ENCE, I define a Z-Variance Error
(ZVE) as the mean deviation from 1 of the binned $v_{i}$ statistics
\begin{equation}
\mathrm{ZVE}=\exp\left(\frac{1}{N}\sum_{i=1}^{N}|\ln v_{i}|\right)
\end{equation}
accounting for the fact that the $v_{i}$ are dimensionless scale
statistics. The ZVE is therefore also a MAD-based statistic, but is
should provide a more reliable assessment of local calibration, notably
for schemes with large bins. It is compared to the ENCE in the following
section.

\subsection{Application to ML-UQ datasets}

\noindent The observations made above for a toy model are now tested
on three recent ML-UQ datasets, for both the ENCE and ZVE statistics.
In all cases, the number $N$ of equal-size bins is varied between
1 and a maximum value chosen to avoid bins with less than 30 points.

\subsubsection{Case BUS2022}

\noindent This dataset has been presented in Busk \emph{et al.}\citep{Busk2022}
and reused in a previous study\citep{Pernot2023_Arxiv}. The QM9 validation
dataset consists of M =13\,885 predicted atomization energies ($V$),
uncertainties ($u_{V}$), and reference values ($R$). In absence
of uncertainty on the reference values, these data are transformed
to\textcolor{violet}{{} }$E=R-V$ and $u=u_{V}$. 

\begin{figure}[t]
\noindent \begin{centering}
\includegraphics[width=0.9\textwidth]{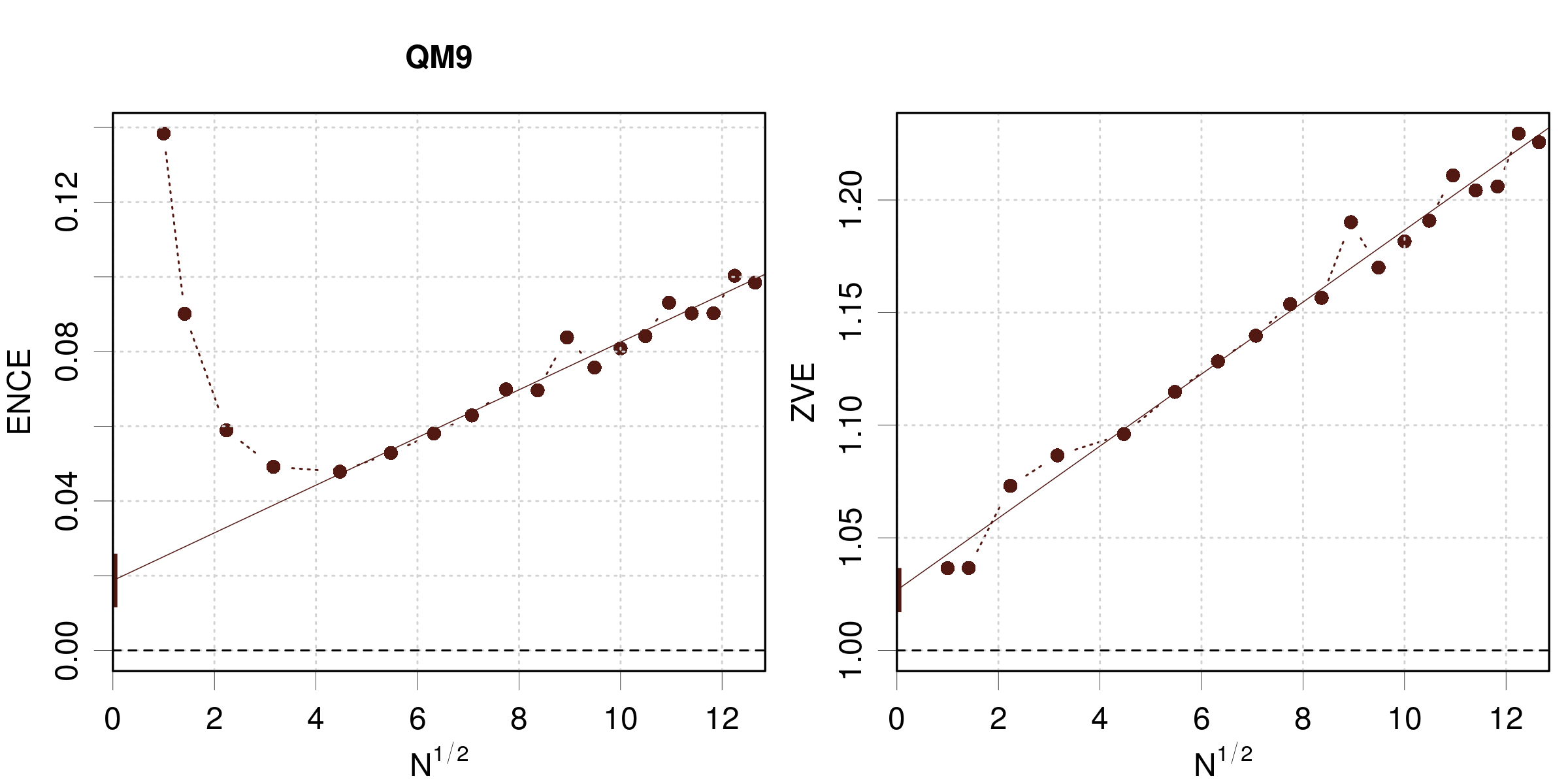}
\par\end{centering}
\caption{\label{fig:Bin-size-dependence-BUS2022}Dependence of the ENCE and
ZVE on the square root of the number for bins $N^{1/2}$ for the BUS2022
QM9 dataset. The linear fits (solid lines) are based on the points
with $N^{1/2}>4$ for the ENCE and all points for the ZVE. The vertical
line segment at the origin represents the 95\% confidence interval
on the fit intercept.}
\end{figure}

The estimation of the ENCE as a function of the number of bins is
presented in Fig.\,\ref{fig:Bin-size-dependence-BUS2022}. The curve
presents a rise at small values of $N$ which was not observed for
the toy problem. In fact, this dataset presents several outstanding
errors at large uncertainties, which produce a large ENCE contribution
into a single bin. The impact of this outstanding bin statistic gets
averaged out as the number of bins increases. Above about 20 bins,
the ENCE rejoins a linear trend with $N^{1/2}$. Busk \emph{et al.}\citep{Busk2022}
report a value of \textasciitilde{}5\,\% for 15 bins, consistent
with our estimation for $N^{1/2}=4$. The linear fit of the values
above $N^{1/2}=4$ provides an intercept value of \textasciitilde{}0.019(3)
(Table\,\ref{tab:Coefficients-of-the}), i.e. a \textasciitilde{}2\,\%
residual relative calibration error which is significantly different
from 0. 
 
\begin{table}[t]
\noindent \begin{centering}
\begin{tabular}{llr@{\extracolsep{0pt}.}lr@{\extracolsep{0pt}.}l}
\hline 
Statistic & Set  & \multicolumn{2}{c}{Intercept } & \multicolumn{2}{c}{Slope}\tabularnewline
\hline 
ENCE & QM9  & 0&019(3)  & 0&0064(3)\tabularnewline
 & Diffusion  & 0&06(1)  & 0&013(2)\tabularnewline
 & Perovskite  & 0&071(6)  & 0&0149(8)\tabularnewline
\hline 
ZVE & QM9  & 1&027(4)  & 0&0160(5)\tabularnewline
 & Diffusion  & 1&11(4)  & 0&039(6)\tabularnewline
 & Perovskite  & 1&11(2)  & 0&053(2)\tabularnewline
\hline 
\end{tabular}
\par\end{centering}
\caption{\label{tab:Coefficients-of-the}Coefficients of the linear fits with
respect to $N^{1/2}$ for the ENCE and ZPE statistics of the three
studied datasets.}
\end{table}

In contrast, the ZVE statistic is not notably impaired by the outliers
and displays a consistent linear trend with $N^{1/2}$ over the whole
range. The linear fit using all points provides an intercept of \textasciitilde{}1.027(4)
(Table\,\ref{tab:Coefficients-of-the}), i.e. a \textasciitilde{}2.7\,\%
relative error on variance calibration, significantly different from
0. 

Both calibration statistics therefore lead us to conclude that the
dataset has small residual calibration errors, but at a level that
is even smaller than the 5\,\% that the original authors considered
as a ``low ENCE''\citep{Busk2022}.

\subsubsection{Case PAL2022}

\noindent The data have been gathered from the supplementary information
of a recent article by Palmer \emph{et al}.\citep{Palmer2022} and
previously reanalyzed\citep{Pernot2023_Arxiv}. I retain here two
sets of errors and uncertainties \emph{before} and \emph{after} calibration
by a bootstrapping method and resulting from the application of a
random forest regression to two materials datasets (Diffusion, $M=2040$
and Perovskites, $M=3836$). The reader is referred to the original
article for more details on the methods and datasets. 

\begin{figure}[t]
\noindent \begin{centering}
\includegraphics[width=0.9\textwidth]{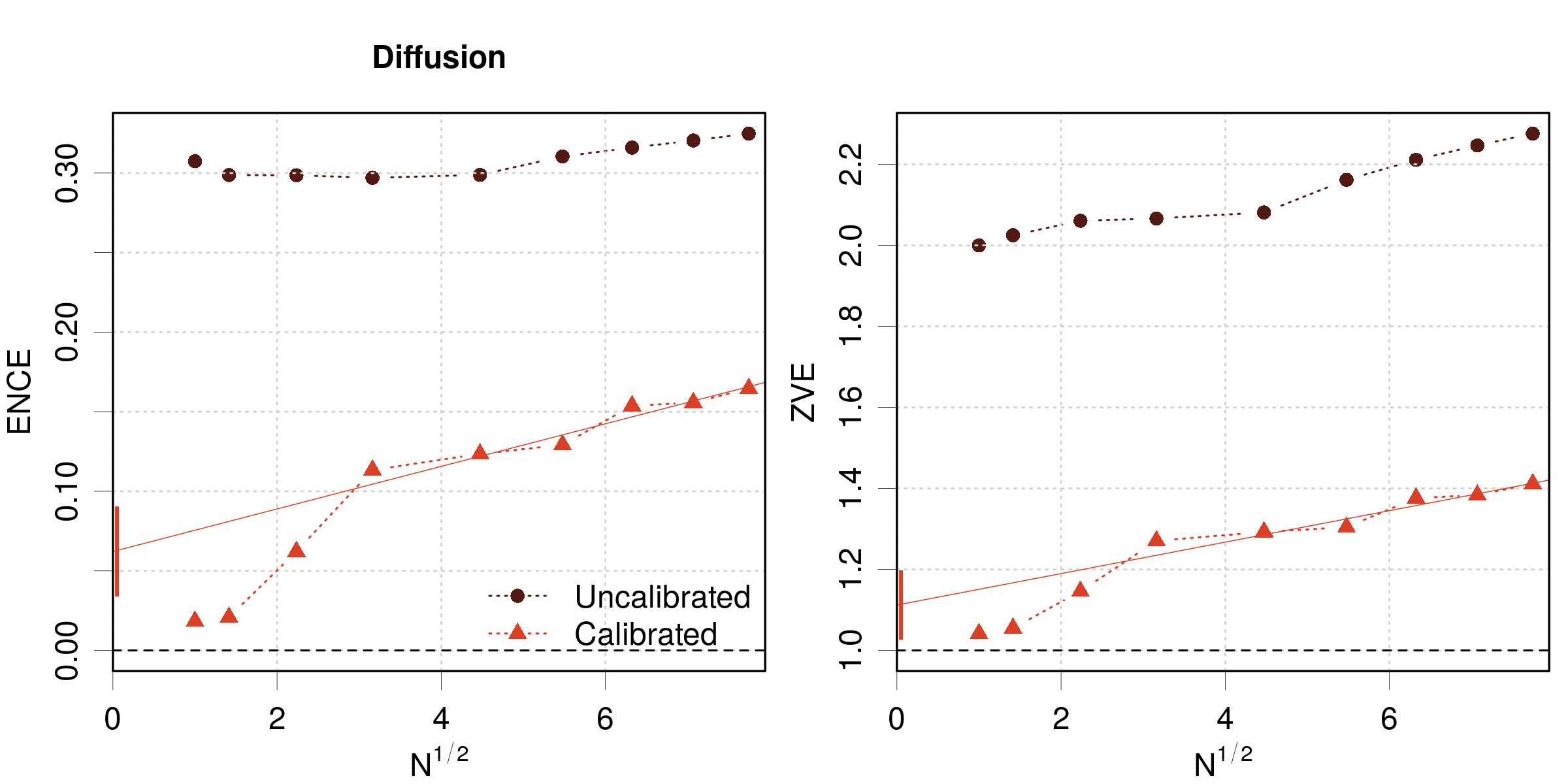}
\par\end{centering}
\noindent \begin{centering}
\includegraphics[width=0.9\textwidth]{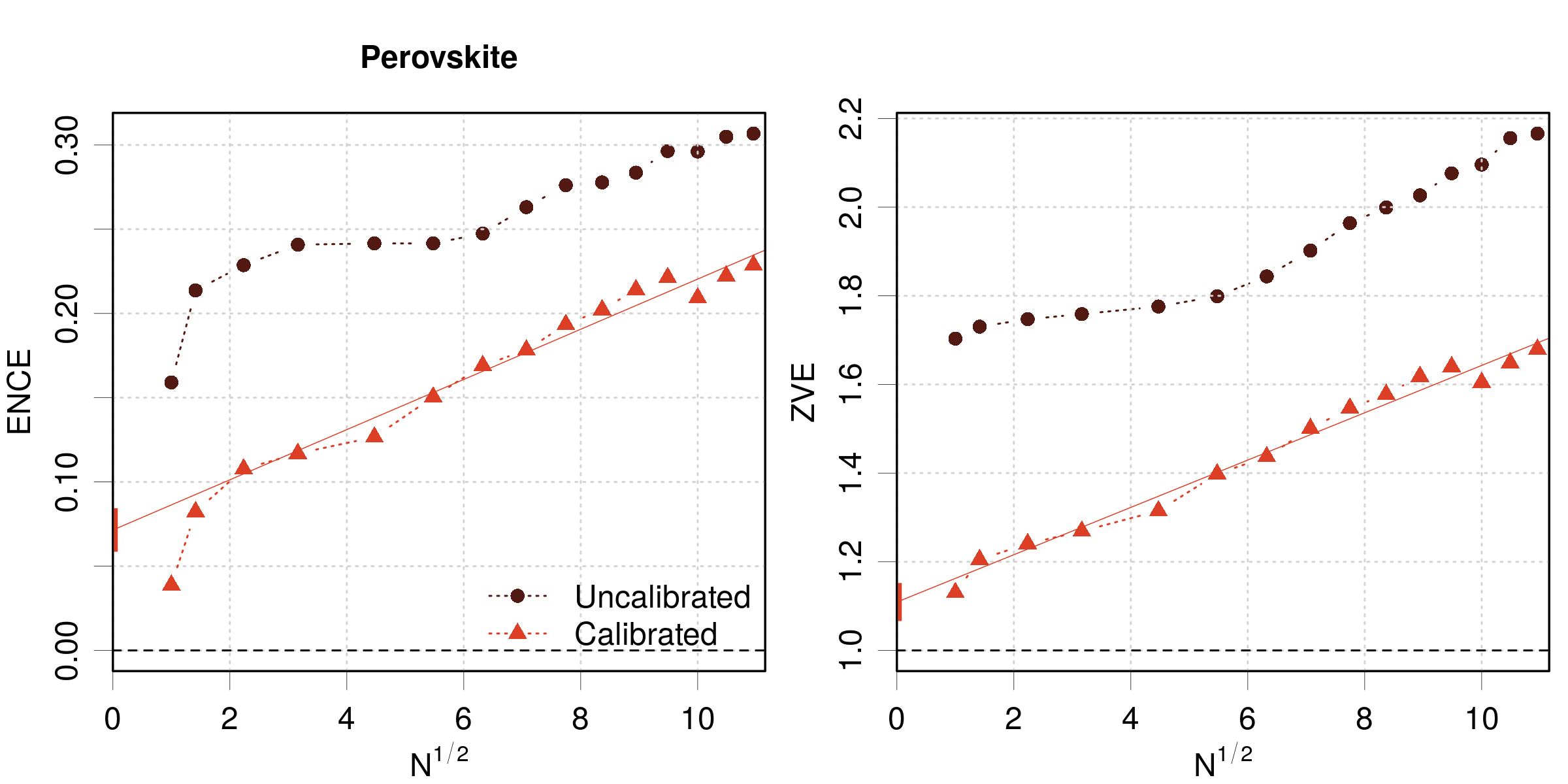}
\par\end{centering}
\caption{\label{fig:PAL2022-Diffusion}Dependence of the ENCE and ZVE on the
square root of the number for bins $N^{1/2}$ for the PAL2022 Diffusion
and Perovskite dataset. The linear fits of the calibrated data (dashed
lines) are based on the points with $N^{1/2}>4$ for Diffusion and
$N^{1/2}>2$ for Perovskite. The vertical line segment at the origin
represents the 95\% confidence interval on the linear fit intercept.}
\end{figure}

The dependence of the ENCE and ZVE with $N^{1/2}$ are presented in
Fig.\,\ref{fig:PAL2022-Diffusion}. Let us first consider the \emph{uncalibrated}
datasets (black dots). Although the dependence on $N^{1/2}$ is sensible,
there is no doubt that the ENCE is about 30\,\% for the Diffusion
data. Considering the shape of the curve with a fall-off below $N^{1/2}=3$,
the ENCE value is more difficult to appreciate for the Perovskite
data, but lies probably around 25\,\%. The ZVE curves for both datasets
present a transition around 4-6 from a more or less constant curve
to a linear increase with $N^{1/2}$ . For these data, using $N=15$
seems to be a reasonable choice as it falls within the more or less
constant part of the curve. 

The curves for the calibrated data (red triangles) also present deviations
from linearity below $N^{1/2}=4$ for Diffusion and $N^{1/2}=2$ for
Perovskite. The corresponding values are excluded from the linear
fit, which leaves only 6 points for the Diffusion dataset and 13 points
for the Perovskite. Nevertheless, the values and confidence intervals
of the intercepts (Table\,\ref{tab:Coefficients-of-the}) lead to
conclude to significant residual calibration errors for both datasets,
between 5 and 10\,\% for the ENCE and about 10\,\% for the ZVE. 

\subsection{Conclusions\label{sec:Discussion-and-conclusion}}

\noindent I have shown through synthetic and ML-UQ datasets that usual
calibration statistics, such as the ENCE and its alternative ZVE,
are strongly sensitive to the choice of the number of bins $N$ for
calibrated or nearly calibrated datasets. Both statistics are globally
linear in $N^{1/2}$, a behavior that can be traced to the use of
the MAD statistic, which is a location estimator for sets with large
bias and a dispersion estimator for unbiased sets. Not treated in
this study, metrics based on maximal deviations are also expected
to be dependent on the binning scheme, in a more trivial way.

A consequence is that comparisons of MAD-based calibration statistics
should always be done on the basis of identical binning schemes. Another
important issue is that interpretation of these calibration statistics
might be problematic when comparing uncalibrated and calibrated datasets,
as they do not convey the extent of residual calibration errors for
the latter. I have shown that the intercept from a linear fit with
respect to $N^{1/2}$ of a calibration statistic for a series of bin
numbers can provide a more reliable estimation of the calibration
error and can also be used as a calibration test.

It appears also that the ZVE is less sensitive than the ENCE to outstanding
values in the datasets, which might make it a more suitable calibration
error statistic.

\subsection*{Acknowledgments}

\noindent I warmly thank J. Busk for providing me the data of the
BUS2022 case.

\subsection*{Author Declarations: Conflict of Interest}

\noindent The author has no conflicts to disclose.

\subsection*{Code and data availability\label{sec:Code-and-data}}

\noindent The code and data to reproduce the results of this article
are available at \url{https://github.com/ppernot/2023_ENCE/releases/tag/v0.0}
and at Zenodo (\url{https://doi.org/10.5281/zenodo.7943886}).\textcolor{orange}{{}
}The \texttt{R},\citep{RTeam2019} \href{https://github.com/ppernot/ErrViewLib}{ErrViewLib}
package implements the \texttt{plotRelDiag} and \texttt{plotLZV} functions
used in the present study to estimate the ENCE and ZVE, respectively,
under version \texttt{ErrViewLib-v1.7.1} (\url{https://github.com/ppernot/ErrViewLib/releases/tag/v1.7.1}),
also available at Zenodo (\url{https://doi.org/10.5281/zenodo.7943906}).

\bibliographystyle{unsrturlPP}
\bibliography{NN}

\begin{thebibliography}{10}

\bibitem{Guo2017}
C.~Guo, G.~Pleiss, Y.~Sun, and K.~Q. Weinberger.
\newblock \href{https://proceedings.mlr.press/v70/guo17a.html}{{On Calibration
  of Modern Neural Networks}}.
\newblock In {\em {International Conference on Machine Learning}}, pages
  1321--1330. 2017.
\newblock URL: \url{https://proceedings.mlr.press/v70/guo17a.html}.

\bibitem{Laves2020}
M.-H. Laves, S.~Ihler, J.~F. Fast, L.~A. Kahrs, and T.~Ortmaier.
\newblock
  \href{https://proceedings.mlr.press/v121/laves20a.html}{Well-calibrated
  regression uncertainty in medical imaging with deep learning}.
\newblock In T.~Arbel, I.~Ben~Ayed, M.~de~Bruijne, M.~Descoteaux, H.~Lombaert,
  and C.~Pal, editors, {\em Proceedings of the Third Conference on Medical
  Imaging with Deep Learning}, volume 121 of {\em Proceedings of Machine
  Learning Research}, pages 393--412. PMLR, 06--08 Jul 2020.
\newblock URL: \url{https://proceedings.mlr.press/v121/laves20a.html}.

\bibitem{Levi2020}
D.~Levi, L.~Gispan, N.~Giladi, and E.~Fetaya.
\newblock \href{http://arxiv.org/abs/1905.11659}{Evaluating and {Calibrating}
  {Uncertainty} {Prediction} in {Regression} {Tasks}}.
\newblock {\em arXiv:1905.11659}, 2020.
\newblock URL: \url{http://arxiv.org/abs/1905.11659}.

\bibitem{Levi2022}
D.~Levi, L.~Gispan, N.~Giladi, and E.~Fetaya.
\newblock \href{http://dx.doi.org/10.3390/s22155540}{{Evaluating and
  Calibrating Uncertainty Prediction in Regression Tasks}}.
\newblock {\em Sensors}, 22(15):5540, 2022.

\bibitem{Scalia2020}
G.~Scalia, C.~A. Grambow, B.~Pernici, Y.-P. Li, and W.~H. Green.
\newblock \href{http://dx.doi.org/10.1021/acs.jcim.9b00975}{Evaluating scalable
  uncertainty estimation methods for deep learning-based molecular property
  prediction}.
\newblock {\em J. Chem. Inf. Model.}, 60:2697--2717, 2020.

\bibitem{Wang2021}
D.~Wang, J.~Yu, L.~Chen, X.~Li, H.~Jiang, K.~Chen, M.~Zheng, and X.~Luo.
\newblock \href{http://dx.doi.org/https://doi.org/10.1186/s13321-021-00551-x}{A
  hybrid framework for improving uncertainty quantification in deep
  learning-based {QSAR} regression modeling}.
\newblock {\em J. Cheminf.}, 13, 2021.

\bibitem{Busk2022}
J.~Busk, P.~B. J{\o}rgensen, A.~Bhowmik, M.~N. Schmidt, O.~Winther, and
  T.~Vegge.
\newblock \href{http://dx.doi.org/10.1088/2632-2153/ac3eb3}{Calibrated
  uncertainty for molecular property prediction using ensembles of message
  passing neural networks}.
\newblock {\em Mach. Learn.: Sci. Technol.}, 3:015012, 2022.

\bibitem{Vazquez-Salazar2022}
L.~I. Vazquez-Salazar, E.~D. Boittier, and M.~Meuwly.
\newblock \href{http://dx.doi.org/10.48550/arXiv.2207.06916}{{Uncertainty
  quantification for predictions of atomistic neural networks}}.
\newblock {\em arXiv:2207.06916}, July 2022.

\bibitem{Frenkel2023}
L.~Frenkel and J.~Goldberger.
\newblock
  \href{https://www.eng.biu.ac.il/goldbej/files/2023/04/LIor_ISBI_2023.pdf}{Calibration
  of a regression network based on the predictive variancewith applications to
  medical images}.
\newblock 2023.
\newblock URL:
  \url{https://www.eng.biu.ac.il/goldbej/files/2023/04/LIor_ISBI_2023.pdf}.

\bibitem{Pernot2022a}
P.~Pernot.
\newblock \href{http://dx.doi.org/10.1063/5.0084302}{The long road to
  calibrated prediction uncertainty in computational chemistry}.
\newblock {\em J. Chem. Phys.}, 156:114109, 2022.

\bibitem{Pernot2022b}
P.~Pernot.
\newblock \href{http://dx.doi.org/10.1063/5.0109572}{Prediction uncertainty
  validation for computational chemists}.
\newblock {\em J. Chem. Phys.}, 157:144103, 2022.

\bibitem{Pernot2023_Arxiv}
P.~Pernot.
\newblock \href{http://dx.doi.org/10.48550/arXiv.2303.07170}{{Validation of
  uncertainty quantification metrics: a primer based on the consistency and
  adaptivity concepts}}.
\newblock {\em arXiv:2303.07170}, 2023.
\newblock \href {http://arxiv.org/abs/2303.07170} {\path{arXiv:2303.07170}}.

\bibitem{Angelopoulos2021}
A.~N. Angelopoulos and S.~Bates.
\newblock \href{http://dx.doi.org/10.48550/arXiv.2107.07511}{{A Gentle
  Introduction to Conformal Prediction and Distribution-Free Uncertainty
  Quantification}}.
\newblock {\em arXiv:2107.07511}, July 2021.

\bibitem{Palmer2022}
G.~Palmer, S.~Du, A.~Politowicz, J.~P. Emory, X.~Yang, A.~Gautam, G.~Gupta,
  Z.~Li, R.~Jacobs, and D.~Morgan.
\newblock \href{http://dx.doi.org/10.1038/s41524-022-00794-8}{{Calibration
  after bootstrap for accurate uncertainty quantification in regression
  models}}.
\newblock {\em npj Comput. Mater.}, 8:1--9, 2022.

\bibitem{Watts2022}
S.~Watts and L.~Crow.
\newblock \href{http://dx.doi.org/10.48550/arXiv.2210.02848}{{The Shannon
  Entropy of a Histogram}}.
\newblock {\em arXiv:2210.02848}, October 2022.

\bibitem{Pernot2018}
P.~Pernot and A.~Savin.
\newblock \href{http://dx.doi.org/10.1063/1.5016248}{Probabilistic performance
  estimators for computational chemistry methods: the empirical cumulative
  distribution function of absolute errors}.
\newblock {\em J. Chem. Phys.}, 148:241707, 2018.

\bibitem{RTeam2019}
{R Core Team}.
\newblock \href{http://www.R-project.org/}{{\em {R}: {A} {L}anguage and
  {E}nvironment for {S}tatistical {C}omputing}}.
\newblock R Foundation for Statistical Computing, Vienna, Austria, 2019.
\newblock URL: \url{http://www.R-project.org/}.

\end{thebibliography}

\appendix

\subsection*{APPENDIX}

\subsection{MAD of binned statistics\label{subsec:MAD-of-binned}}

\noindent Let us consider a set of M draws from a zero-centered (unbiased)
normal distribution $X\sim N(0,u_{X})$. These data are binned into
$N$ equal-size bins of size $k$ (assuming for convenience that $M$
is always divisible by $N)$. The average of $X$ is estimated within
each bin
\begin{equation}
\mu_{j}=\frac{1}{k}\sum_{i\in B_{j}}X_{i}
\end{equation}
and the uncertainty on the mean is 
\begin{equation}
u_{j}\simeq\frac{u_{X}}{\sqrt{k}}=u_{X}\sqrt{\frac{N}{M}}
\end{equation}
from which one estimates the MAD of $\mu_{j}$
\begin{equation}
m=\frac{1}{N}\sum_{j=1}^{N}|\mu_{j}|
\end{equation}
As $\mu_{j}$ has a zero-centered normal distribution, $|\mu_{i}|$
has a half-normal distribution with mean $\overline{\mu_{j}}=u_{j}\sqrt{2/\pi}$,
making the link of the MAD with a dispersion statistic. Finally, the
mean value for the MAD is
\begin{align}
m & =u_{X}\sqrt{\frac{2}{\pi M}}\sqrt{N}
\end{align}
Therefore, $m$ is proportional to $N^{1/2}$, showing the direct
impact of the binning strategy on the value of $m$. 
\end{document}